\documentclass[10pt,twocolumn,letterpaper]{article}

\usepackage{iccv}
\usepackage{times}
\usepackage{epsfig}
\usepackage{graphicx}
\usepackage{subfig}
\usepackage{amsmath}
\usepackage{amssymb}
\usepackage[flushleft]{threeparttable} 
\usepackage{multirow}
\usepackage{tabularx}
\usepackage{booktabs}
\usepackage{footnote}
\usepackage{xcolor}
\makesavenoteenv{tabularx}
\makesavenoteenv{tabular}
\makesavenoteenv{table}

\usepackage[pagebackref=true,breaklinks=true,letterpaper=true,colorlinks,bookmarks=false]{hyperref}

\iccvfinalcopy % *** Uncomment this line for the final submission

\def\iccvPaperID{5446} % *** Enter the ICCV Paper ID here

\ificcvfinal\fi
\begin{document}

%%%%%%%%% TITLE
\title{Recurrent Embedding Aggregation Network for Video Face Recognition}

% \author{
%     Sixue Gong\IEEEauthorrefmark{1},\;
%     Yichun Shi\IEEEauthorrefmark{1},\;
%     Anil K. Jain\IEEEauthorrefmark{1},\;
%     Nathan D. Kalka\IEEEauthorrefmark{2}\\
%     \IEEEauthorrefmark{1}Michigan State University, East Lansing, MI\\
%     \IEEEauthorrefmark{2}Noblis, Bridgeport, WV\\
% {\tt\small gongsixu@msu.edu, shiyichu@msu.edu, jain@cse.msu.edu, nathan.kalka@noblis.org }
% }
\author{
    Sixue Gong,\;
    Yichun Shi,\;
    Anil K. Jain\\
    Michigan State University, East Lansing, MI\\
{\tt\small gongsixu@msu.edu, shiyichu@msu.edu, jain@cse.msu.edu }
}

\maketitle
%\thispagestyle{empty}

%%%%%%%%% ABSTRACT
\begin{abstract}
Recurrent networks have been successful in analyzing temporal data and have been widely used for video analysis. However, for video face recognition, where the base CNNs trained on large-scale data already provide discriminative features, using Long Short-Term Memory (LSTM), a popular recurrent network, for feature learning could lead to overfitting and degrade the performance instead. We propose a Recurrent Embedding Aggregation Network (REAN) for set to set face recognition. Compared with LSTM, REAN is robust against overfitting because it only learns how to aggregate the pre-trained embeddings rather than learning representations from scratch. Compared with quality-aware aggregation methods, REAN can take advantage of the context information to circumvent the noise introduced by redundant video frames. Empirical results on three public domain video face recognition datasets, IJB-S, YTF, and PaSC show that the proposed REAN significantly outperforms naive CNN-LSTM structure and quality-aware aggregation methods.
\end{abstract}

%%%%%%%%% BODY TEXT
\section{Introduction}

% 1. Why is video face recognition an important problem?\\
An increasing number of videos captured by both mobile devices and CCTV systems around the world\footnote{Close to 200 million surveillance cameras have already been installed across China, which amounts to approximately 1 camera per 7 citizens. Approximately 40 million surveillance cameras were active in the United States in 2014, which amounts to approximately 1 camera per 8 citizens~\cite{CCTV_CN}.} has generated an urgent need for robust and accurate face recognition in video. Face recognition approaches for high quality still images (controlled capture and cooperative subjects) are not able to deal with challenges in face recognition in unconstrained videos. On the other hand, a video sequence contains more information about the subject in terms of temporal context (varying pose, expression, and motion) than still images. Thus, one of the key challenges in video face recognition is how to effectively combine facial information available across multiple video frames to improve face recognition accuracy.

\begin{figure}[t]
    \centering
    \captionsetup{font=footnotesize}
    \subfloat[LSTM]{\centering\includegraphics[width=0.48\linewidth]{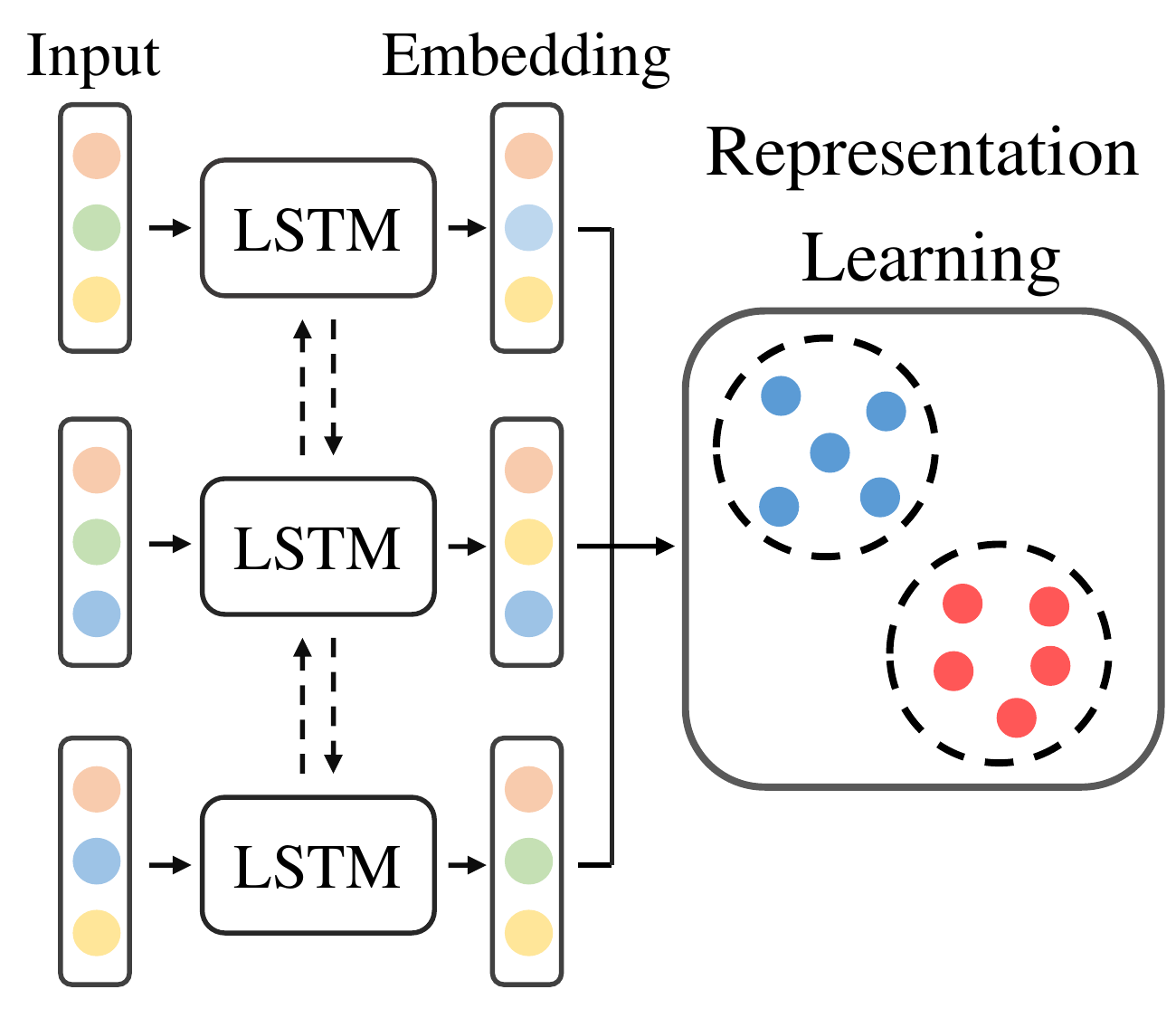}}\hfill
    \subfloat[REAN]{\centering\includegraphics[width=0.48\linewidth]{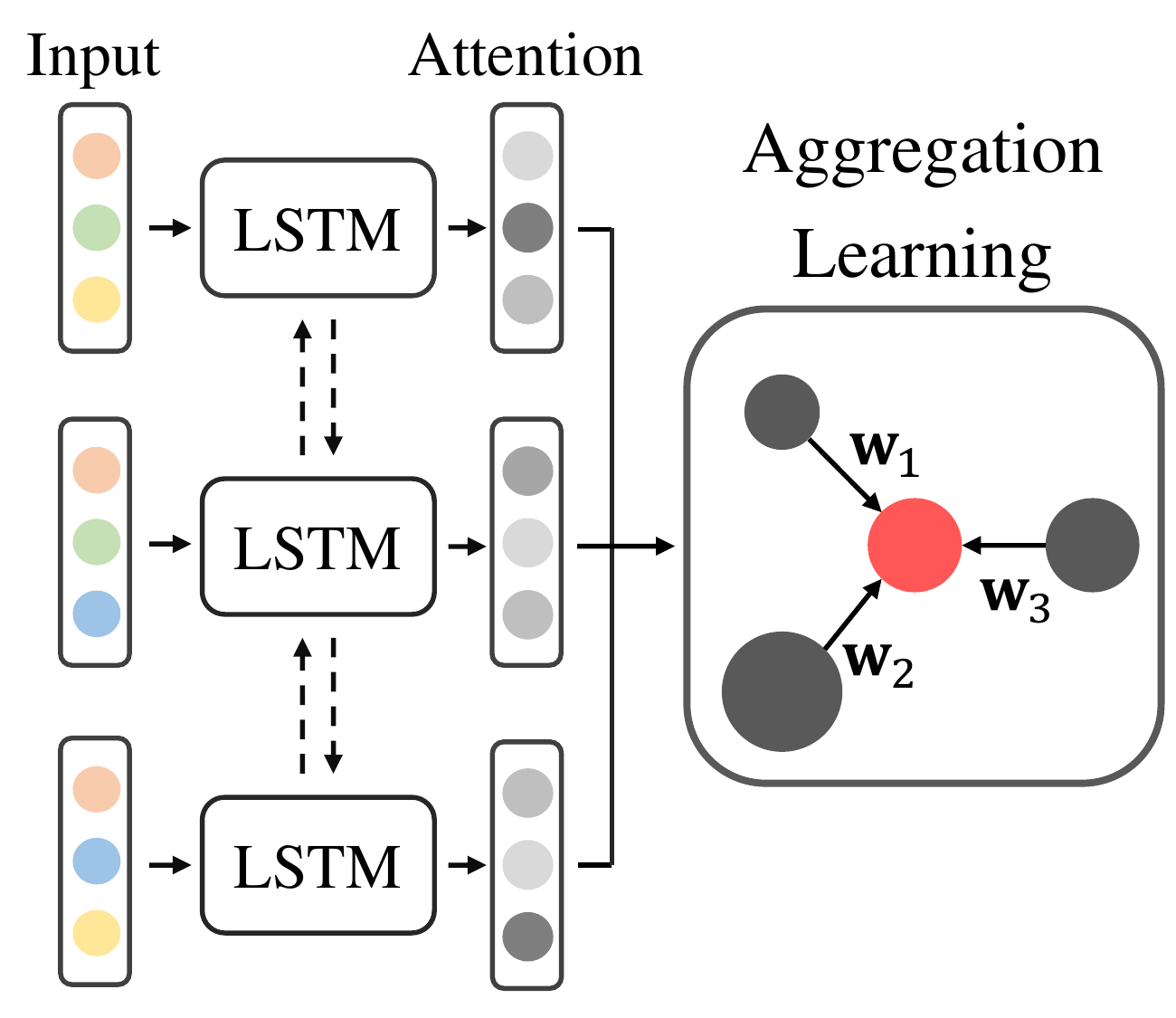}}
    % \subfloat[Quality-aware Pooling\\\centering(with redundant inputs)]{\includegraphics[width=0.46\linewidth]{figs/contextaware3.pdf}}\hfill
    % \subfloat[Context-aware Pooling]{\includegraphics[width=0.46\linewidth]{figs/contextaware4.pdf}}\hfill
    \vspace{-0.75em}\caption{Difference between naive LSTM and the proposed Recurrent Embedding Aggregation Network (REAN). LSTM needs to learn new representations from scratch and hence could easily overfit. In comparison, REAN utilizes pre-trained embeddings and integrates them using context information.}\vspace{-1.0em}
    \label{fig:frontpage}
\end{figure}

\begin{figure*}
\captionsetup{font=small}
\footnotesize
    \centering
    \newcommand{\vshrink}{\vspace{-10px}}
    \begin{minipage}{0.24\linewidth}
    \includegraphics[width=0.25\linewidth]{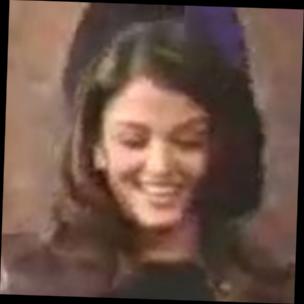}\hfill
    \includegraphics[width=0.25\linewidth]{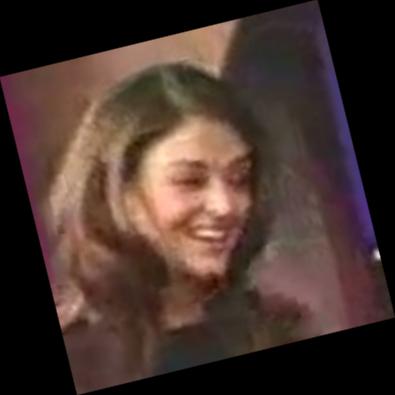}\hfill
    \includegraphics[width=0.25\linewidth]{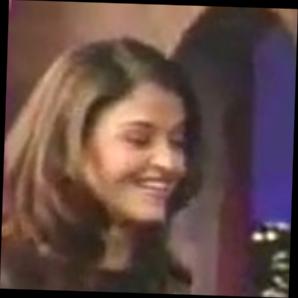}\hfill
    \includegraphics[width=0.25\linewidth]{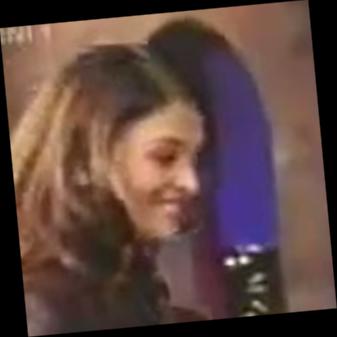}\\
    \includegraphics[width=0.25\linewidth]{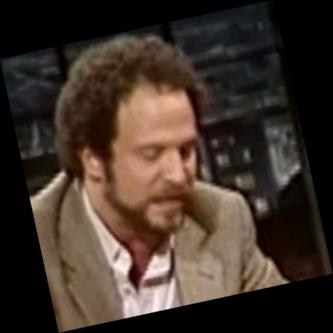}\hfill
    \includegraphics[width=0.25\linewidth]{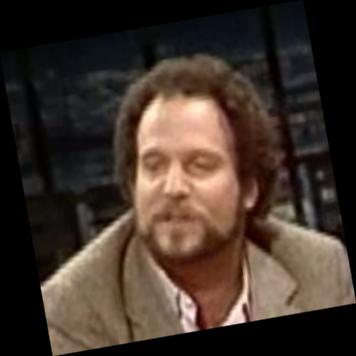}\hfill
    \includegraphics[width=0.25\linewidth]{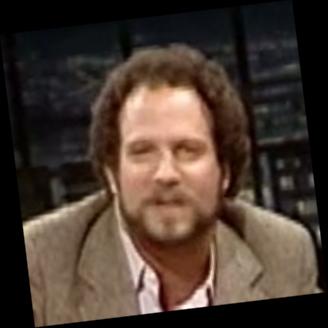}\hfill
    \includegraphics[width=0.25\linewidth]{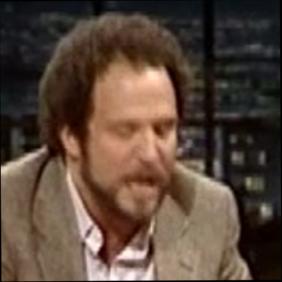}\\
    \includegraphics[width=0.25\linewidth]{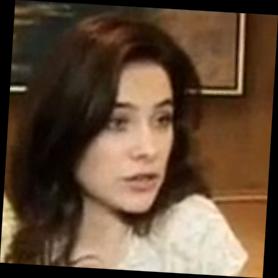}\hfill
    \includegraphics[width=0.25\linewidth]{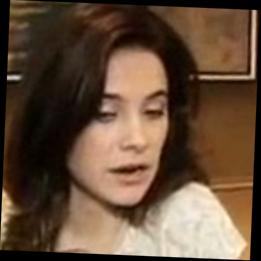}\hfill
    \includegraphics[width=0.25\linewidth]{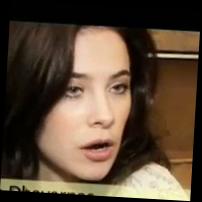}\hfill
    \includegraphics[width=0.25\linewidth]{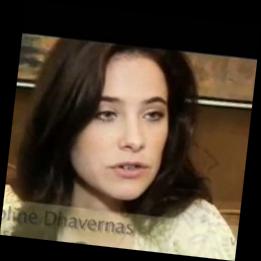}\\
    \vspace{-1.8em}\begin{center} (a) \label{fig:img_ytf}YoutubeFaces \cite{wolf2011face}\end{center}
    \end{minipage}\hfill
    \begin{minipage}{0.24\linewidth}
    \includegraphics[width=0.25\linewidth]{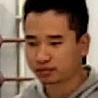}\hfill
    \includegraphics[width=0.25\linewidth]{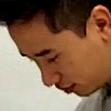}\hfill
    \includegraphics[width=0.25\linewidth]{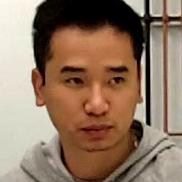}\hfill
    \includegraphics[width=0.25\linewidth]{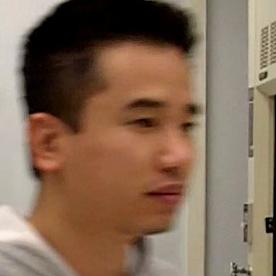}\\
    \includegraphics[width=0.25\linewidth]{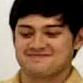}\hfill
    \includegraphics[width=0.25\linewidth]{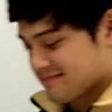}\hfill
    \includegraphics[width=0.25\linewidth]{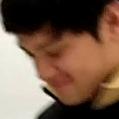}\hfill
    \includegraphics[width=0.25\linewidth]{figs/dataset/pasc/06045d446-160.jpg}\\
    \includegraphics[width=0.25\linewidth]{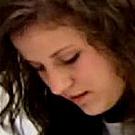}\hfill
    \includegraphics[width=0.25\linewidth]{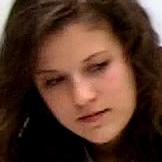}\hfill
    \includegraphics[width=0.25\linewidth]{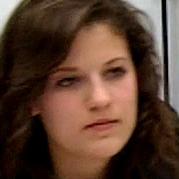}\hfill
    \includegraphics[width=0.25\linewidth]{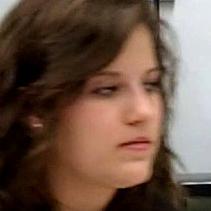}\\
    \vspace{-1.8em}\begin{center} (b) \label{fig:img_pasc} PaSC \cite{beveridge2013challenge}\end{center}
    \end{minipage}\hfill
    \begin{minipage}{0.24\linewidth}
    \includegraphics[width=0.25\linewidth]{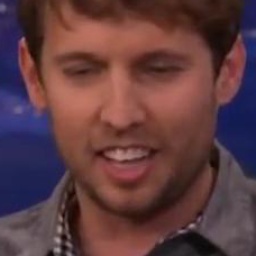}\hfill
    \includegraphics[width=0.25\linewidth]{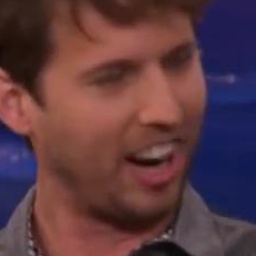}\hfill
    \includegraphics[width=0.25\linewidth]{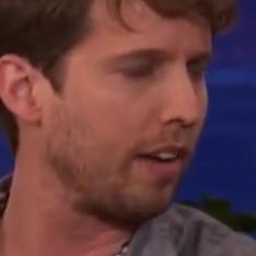}\hfill
    \includegraphics[width=0.25\linewidth]{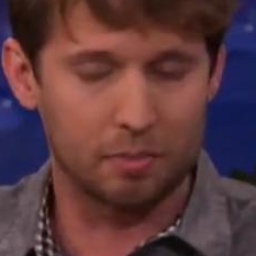}\\
    \includegraphics[width=0.25\linewidth]{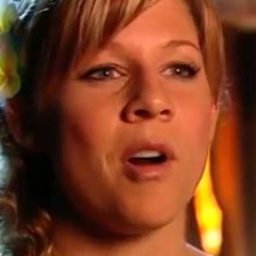}\hfill
    \includegraphics[width=0.25\linewidth]{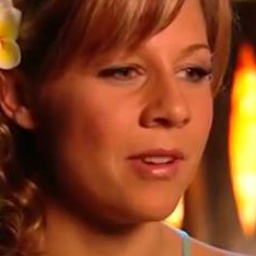}\hfill
    \includegraphics[width=0.25\linewidth]{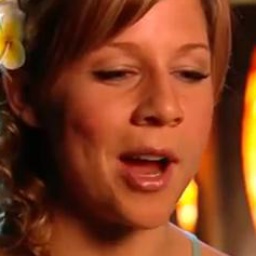}\hfill
    \includegraphics[width=0.25\linewidth]{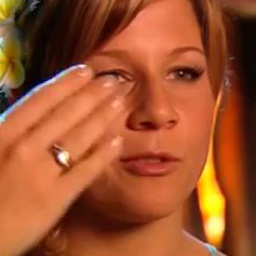}\\
    \includegraphics[width=0.25\linewidth]{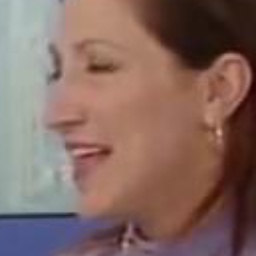}\hfill
    \includegraphics[width=0.25\linewidth]{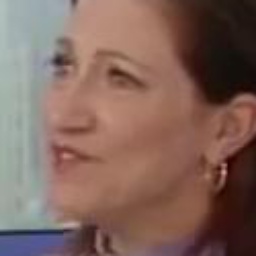}\hfill
    \includegraphics[width=0.25\linewidth]{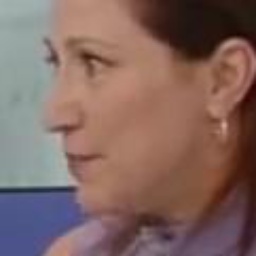}\hfill
    \includegraphics[width=0.25\linewidth]{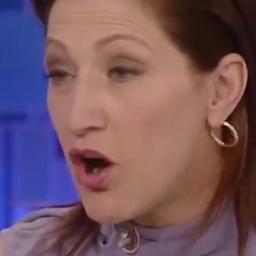}\\
    \vspace{-1.8em}\begin{center} (c) \label{fig:img_ijba}UMDFaceVideo \cite{klare2015pushing}\end{center}
    \end{minipage}\hfill
    \begin{minipage}{0.24\linewidth}
    \includegraphics[width=0.25\linewidth]{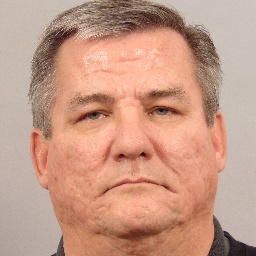}\hfill
    \includegraphics[width=0.25\linewidth]{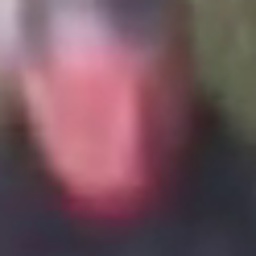}\hfill
    \includegraphics[width=0.25\linewidth]{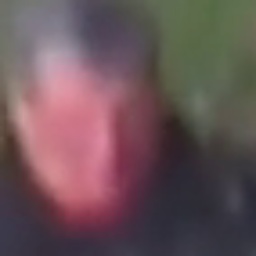}\hfill
    \includegraphics[width=0.25\linewidth]{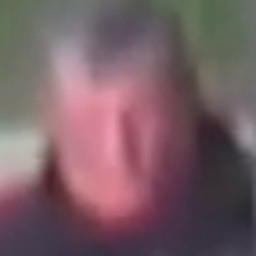}\\
    \includegraphics[width=0.25\linewidth]{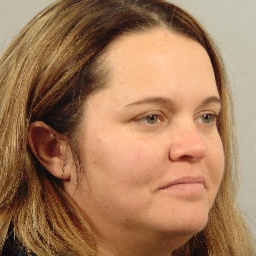}\hfill
    \includegraphics[width=0.25\linewidth]{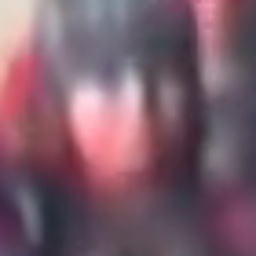}\hfill
    \includegraphics[width=0.25\linewidth]{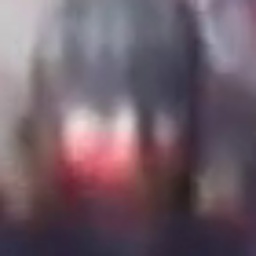}\hfill
    \includegraphics[width=0.25\linewidth]{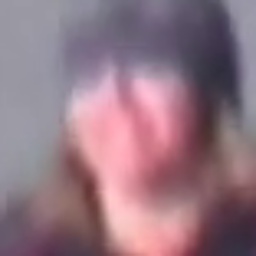}\\
    \includegraphics[width=0.25\linewidth]{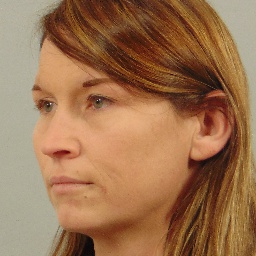}\hfill
    \includegraphics[width=0.25\linewidth]{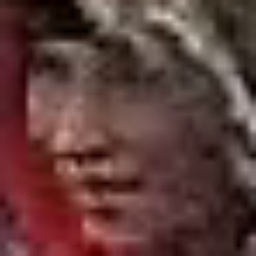}\hfill
    \includegraphics[width=0.25\linewidth]{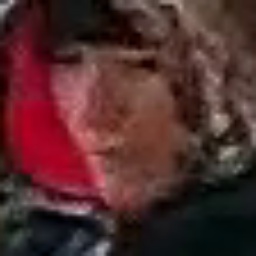}\hfill
    \includegraphics[width=0.25\linewidth]{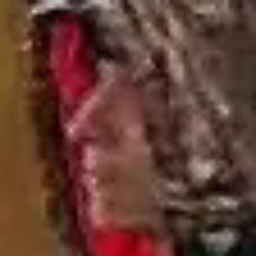}\\
    \vspace{-1.8em}\begin{center} (d) \label{fig:img_ijbs}IJB-S \cite{Kalka2018IJBS}\end{center}
    \end{minipage}\\
    \vspace{-0.5em}\caption{Example images from different video datasets. YoutubeFace, PaSC and UMDFaceVideo contain only video frames. IJB-S includes both still images and videos. The first column of IJB-S shows still images, followed by video frames of the respective subjects in the next three columns.}\vspace{-0.5em}
    \label{fig:dataset}
\end{figure*}
% 2. What are challenges in video face recognition?\\

Deep Neural Networks (DNNs) have shown the ability to learn face representations that are robust to occlusions, image blur and large pose variations to achieve high recognition performance on semi-constrained still face recognition benchmarks~\cite{Schroff_2015_CVPR,taigman2014deepface,liu2017sphereface,sun2015deeply}. While face recognition in surveillance video and unconstrained still face images share similar challenges, video sequences from CCTV are generally of low resolution and may contain noisy frames with poor quality and unfavorable viewing angles (See Figures \textcolor{red}{2} (d) and~\ref{fig:ijbs-s}). Such noisy frames will undermine the overall performance of video face recognition if we directly use recognition methods developed for still images.

% 3. How do state-of-the-art work resolve this problem? (spatial-temporal information)\\
State-of-the-art methods for face recognition in video represent a subject's face as an unordered set of vector and the recognition is posed as estimating the similarity between face templates \cite{arandjelovic2005face, cevikalp2010face, harandi2011graph, wang2008manifold, zhao2019multi}. However, this is not computationally efficient as one needs to compare similarities on all feature vectors between two face templates. Thus, it is preferable to aggregate feature vectors into a compact feature vector for each template \cite{yang2017neural, zhong2018ghostvlad, liu2018dependency, xie2018comparator, rao2017learning, sohn2017unsupervised, rao2017attention, liu2017quality}. Most methods aggregate image sets based on the quality of each image or video frame but ignores the contextual information.

% 4. How do we resolve this problem? Contributions.\\
In this paper, we propose a \textit{Recurrent Embedding Aggregation Network (REAN)} for video face recognition. LSTM models have been shown to robustly grasp information from sequential data by introducing a memory mechanism to utilize the context information. In spite of this, directly using a LSTM for learning video face representations could lead to degraded performance by dropping the discriminative features learned by embedding CNNs. Instead, the proposed REAN learns to aggregate the embeddings of face images by leveraging the context-awareness of LSTMs. Instead of a feature vector for matching, the LSTM in our case outputs an attention vector, which can be used for integration.
% Therefore, we propose REAN to predict an adaptive quality vector for each face feature vector based on the context information in a template, and all the feature vectors in the template are then aggregated with the corresponding quality vectors into a single vector as the template representation. Since the network sequentially observes related image vectors in a template before it makes predictions, the output quality vectors are not only based on the current input image but all the previous inputs as well. Thus, the quality score learned by REAN is not an absolute value of the current image, but a relative score that reflects its quality compared to the other images in the template. 
Experimental results on IJB-S dataset~\cite{Kalka2018IJBS} and other template/video matching benchmarks show that the proposed REAN significantly improves the face recognition performance compared with average pooling and other state-of-the-art aggregation methods. Specific contributions of the paper are listed below:
\begin{itemize}
    \item A Recurrent Embedding Aggregation Network (REAN) that aggregates deep feature vectors based on quality and context information, resulting in better representation for video face recognition. \vspace{-5px}
    \item The attention scores of one video frame provided by REAN present the relative discrimination power given the other frames in the video. \vspace{-5px}
    \item REAN achieves state-of-the-art performance on a surveillance benchmark IJB-S \cite{Kalka2018IJBS} and two other face recognition benchmarks, YouTube Faces \cite{wolf2011face}, and PaSC \cite{beveridge2013challenge}.
\end{itemize}

% \begin{figure}[t]
%     \centering
%     \captionsetup{font=footnotesize}
%     \subfloat[Quality-aware Attention]{\centering\includegraphics[width=0.5\linewidth]{figs/lstm_front_new_1.pdf}}\hfill
%     \subfloat[Context-aware Attention]{\centering\includegraphics[width=0.5\linewidth]{figs/lstm_front_new_2.pdf}}
%     % \subfloat[Quality-aware Pooling\\\centering(with redundant inputs)]{\includegraphics[width=0.46\linewidth]{figs/contextaware3.pdf}}\hfill
%     % \subfloat[Context-aware Pooling]{\includegraphics[width=0.46\linewidth]{figs/contextaware4.pdf}}\hfill
%     \vspace{-0.75em}\caption{Difference between quality-aware attention and context-aware attention. Quality-aware attention could still be biased towards low-quality images because of redundancy while context-aware attention dynamically decides the attention for feature aggregation by considering the context information in the video.}\vspace{-1.0em}
%     \label{fig:frontpage}
% \end{figure}

%-------------------------------------------------------------------------
\section{Related Work}

\subsection{Facial Analysis with RNN}
Many existing approaches for facial analysis of videos have utilized RNNs to account for the temporal dependencies in sequences of frames. For example, Gu~\etal~\cite{gu2017dynamic} proposed an end-to-end RNN-based approach to head pose estimation and facial landmark estimation in videos. As for recognition tasks, Ren~\etal~\cite{ren2009pose} attempted to address large out-of-plane pose invariant face recognition in image sequences by using a Cellular Simultaneous Recurrent Network (CSRN). Graves~\etal~\cite{graves2008facial} employed RNN that accepts a sequence of Candide face features as input for facial expression recognition. RNN has also been widely used for face emotion recognition~\cite{zhang2018spatial, fan2016video, ebrahimi2015recurrent} from videos.

\subsection{Video Face Recognition}
State-of-the-art methods for video face recognition can primarily be summarized into three categories: \textit{space}-model, \textit{classifier}-model, and \textit{aggregation}-model. Many traditional \textit{space}-models attempt to estimate a feature space where all the video frames can be embedded. Such feature space can be represented as probabilistic distribution~\cite{shakhnarovich2002face,arandjelovic2005face}, n-order statistics~\cite{lu2013image}, affine hulls~\cite{hu2011sparse,cevikalp2010face,yang2013face}, SPD matrices~\cite{huang2015log}, and manifolds~\cite{lee2003video,harandi2011graph,wang2008manifold}. \textit{classifier}-models~\cite{wolf2011face,parchami2017using} are proposed to learn face representations based on videos or image sets;  \textit{aggregation}-models strive to fuse the identity-relevant information in the face templates/videos to attain both efficiency and recognition accuracy. Best-Rowden~\etal~\cite{best2014unconstrained} showed that combining multiple sources of face media~\footnote{Face media refers to a collection of sources of face information, for example, video tracks, multiple still images, 3D face models, verbal descriptions and face sketches.} boosts the recognition performance for identifying a person of interest. With the ubiquity of deep learning algorithms for face recognition, most recent methods aim to aggregate a set of deep feature vectors into a single vector. Compared to simply averaging all vectors~\cite{ding2018trunk,chen2018unconstrained}, fusing features with the associated visual quality shows more promising results in recognizing faces in unconstrained videos. Ranjan~\etal~\cite{ranjan2018crystal} utilized face detection scores as measures of face quality to rescale the face similarity scores. 
%Another fusion idea is to predict quality values for deep features by learning from data. 
Yang~\etal~\cite{yang2017neural} and Liu~\etal~\cite{liu2017quality} proposed to use an additional network module that predicts a quality score for each feature vector and aggregates the vectors with the assigned scores. Gong~\etal~\cite{gong2019video} extended the aggregation model by considering component-wise quality prediction
%: each component of feature vectors is aggregated independently with the corresponding quality weight. 
Rao~\etal~\cite{rao2017attention} used LSTM to learn temporal features while use reinforcement learning to drop the features of low-quality images. \cite{liu2018dependency} proposed a dependency-aware pooling by modeling the relationship of images within a set and using reinforcement learning for image quality prediction. However, none of these approaches have address the redundancy issue in the video frames.

%-------------------------------------------------------------------------
\section{Motivation}

\subsection{Context Information}

\begin{figure*}
\setlength{\tabcolsep}{0pt}
\captionsetup{font=small}
\footnotesize
    \centering
    \begin{tabularx}{1.0\linewidth}{cccccccccccccccccccc}
    \toprule\\[-1.2em]
    \includegraphics[width=0.05\linewidth]{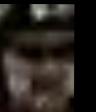} &
    \includegraphics[width=0.05\linewidth]{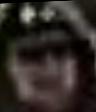} &
    \includegraphics[width=0.05\linewidth]{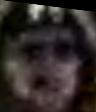} &
    \includegraphics[width=0.05\linewidth]{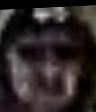} &
    \includegraphics[width=0.05\linewidth]{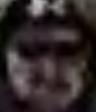} &
    \includegraphics[width=0.05\linewidth]{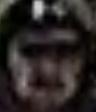} &
    \includegraphics[width=0.05\linewidth]{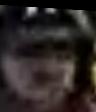} &
    \includegraphics[width=0.05\linewidth]{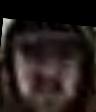} &
    \includegraphics[width=0.05\linewidth]{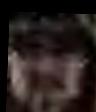} &
    \includegraphics[width=0.05\linewidth]{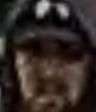} &
    \includegraphics[width=0.05\linewidth]{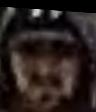} &
    \includegraphics[width=0.05\linewidth]{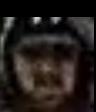} &
    \includegraphics[width=0.05\linewidth]{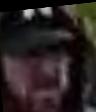} &
    \includegraphics[width=0.05\linewidth]{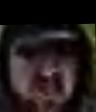} &
    \includegraphics[width=0.05\linewidth]{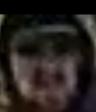} &
    \includegraphics[width=0.05\linewidth]{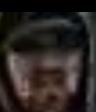} &
    \includegraphics[width=0.05\linewidth]{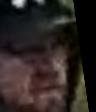} &
    \includegraphics[width=0.05\linewidth]{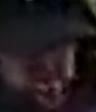} &
    \includegraphics[width=0.05\linewidth]{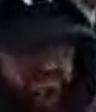} &
    \includegraphics[width=0.05\linewidth]{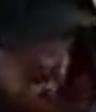} \\[-0.5em]
    \includegraphics[width=0.05\linewidth]{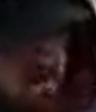} &
    \includegraphics[width=0.05\linewidth]{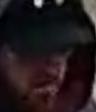} &
    \includegraphics[width=0.05\linewidth]{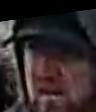} &
    \includegraphics[width=0.05\linewidth]{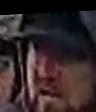} &
    \includegraphics[width=0.05\linewidth]{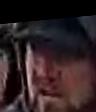} &
    \includegraphics[width=0.05\linewidth]{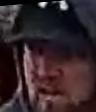} &
    \includegraphics[width=0.05\linewidth]{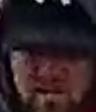} &
    \includegraphics[width=0.05\linewidth]{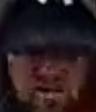} &
    \includegraphics[width=0.05\linewidth]{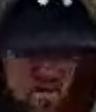} &
    \includegraphics[width=0.05\linewidth]{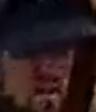} &
    \includegraphics[width=0.05\linewidth]{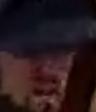} &
    \includegraphics[width=0.05\linewidth]{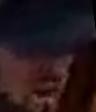} &
    \includegraphics[width=0.05\linewidth]{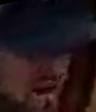} &
    \includegraphics[width=0.05\linewidth]{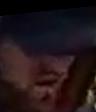} &
    \includegraphics[width=0.05\linewidth]{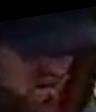} &
    \includegraphics[width=0.05\linewidth]{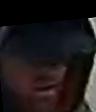} &
    \includegraphics[width=0.05\linewidth]{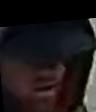} &
    \includegraphics[width=0.05\linewidth]{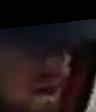} &
    \includegraphics[width=0.05\linewidth]{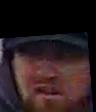} &
    \includegraphics[width=0.05\linewidth]{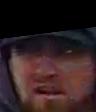} \\[-0.2em]\midrule\\[-1.2em]
    \includegraphics[width=0.05\linewidth]{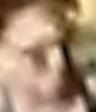} &
    \includegraphics[width=0.05\linewidth]{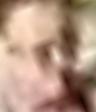} &
    \includegraphics[width=0.05\linewidth]{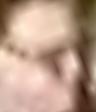} &
    \includegraphics[width=0.05\linewidth]{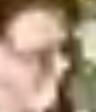} &
    \includegraphics[width=0.05\linewidth]{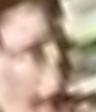} &
    \includegraphics[width=0.05\linewidth]{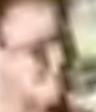} &
    \includegraphics[width=0.05\linewidth]{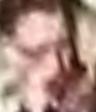} &
    \includegraphics[width=0.05\linewidth]{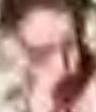} &
    \includegraphics[width=0.05\linewidth]{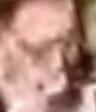} &
    \includegraphics[width=0.05\linewidth]{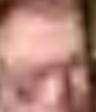} &
    \includegraphics[width=0.05\linewidth]{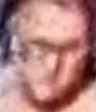} &
    \includegraphics[width=0.05\linewidth]{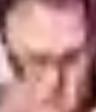} &
    \includegraphics[width=0.05\linewidth]{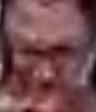} &
    \includegraphics[width=0.05\linewidth]{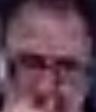} &
    \includegraphics[width=0.05\linewidth]{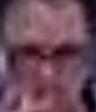} &
    \includegraphics[width=0.05\linewidth]{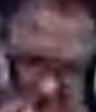} &
    \includegraphics[width=0.05\linewidth]{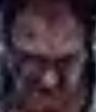} &
    \includegraphics[width=0.05\linewidth]{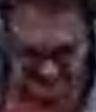} &
    \includegraphics[width=0.05\linewidth]{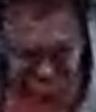} &
    \includegraphics[width=0.05\linewidth]{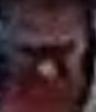} \\[-0.5em]
    \includegraphics[width=0.05\linewidth]{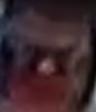} &
    \includegraphics[width=0.05\linewidth]{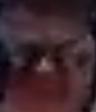} &
    \includegraphics[width=0.05\linewidth]{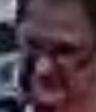} &
    \includegraphics[width=0.05\linewidth]{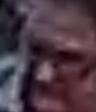} &
    \includegraphics[width=0.05\linewidth]{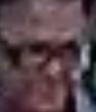} &
    \includegraphics[width=0.05\linewidth]{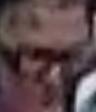} &
    \includegraphics[width=0.05\linewidth]{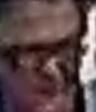} &
    \includegraphics[width=0.05\linewidth]{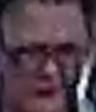} &
    \includegraphics[width=0.05\linewidth]{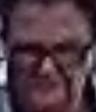} &
    \includegraphics[width=0.05\linewidth]{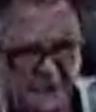} &
    \includegraphics[width=0.05\linewidth]{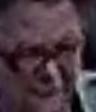} &
    \includegraphics[width=0.05\linewidth]{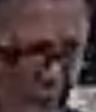} &
    \includegraphics[width=0.05\linewidth]{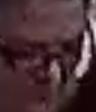} &
    \includegraphics[width=0.05\linewidth]{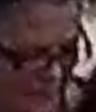} &
    \includegraphics[width=0.05\linewidth]{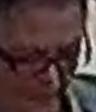} &
    \includegraphics[width=0.05\linewidth]{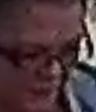} &
    \includegraphics[width=0.05\linewidth]{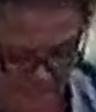} &
    \includegraphics[width=0.05\linewidth]{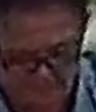} &
    \includegraphics[width=0.05\linewidth]{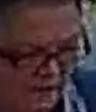} &
    \includegraphics[width=0.05\linewidth]{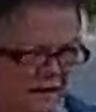} \\[-0.2em]
    \bottomrule
    \end{tabularx}
    \vspace{-0.5em}\caption{Example video frames from IJB-S dataset. We only show videos of two subjects due to the space limit. The first two rows show the frames from one subject and the last rows are frames of another subject.}\vspace{-0.5em}
    %Frames with high quality are highlighted with red bounding boxes.
    \label{fig:ijbs-s}
\end{figure*}

A face video provides a sequence of frames that can be used to perform recognition. This allows us to exploit contextual and temporal information of an identity whose face images are sampled sequentially from a video. For example, a variety of face poses and facial expressions can be present in the trajectory of the face movements in a video. On the other hand, the image quality of video frames obtained from typical CCTV cameras tend to be lower than still images captured under constrained conditions (e.g. passport photos). In addition, video frames may suffer severe motion blur and out-of-focus blur due to camera jitter and small oscillation in the scene. One way to address the large variations in face quality 
%in video streams 
is to select key frames and eliminate poor quality images~\cite{qi2018cnn}. Despite its efficiency, Hassner~\etal~\cite{hassner2016pooling} found that the overall recognition performance is undermined by simply removing low quality images.

Popular formulations of facial information integration in videos or image sets follow the idea of linearly aggregating feature vectors extracted from images in a template by an adaptive weighting scheme to generate a compact face representation for the template~\cite{yang2017neural,liu2017quality,xie2018multicolumn,gong2019video}. A probabilistic model for the face space of noisy embeddings $\mathbf{y}$ extracted from a given facial representation model can be formulated as:
\begin{equation}
    p(\mathbf{y}|\mathbf{I}^*, \mathbf{Y}^*) = \int p(\mathbf{y}|\mathbf{i}, \mathbf{I}^*, \mathbf{Y}^*)p(\mathbf{i}|\mathbf{I}^*, \mathbf{Y}^*)d\mathbf{i},
\end{equation}
where $\mathbf{I}^* = \{\mathbf{i}_1, \mathbf{i}_2, \cdots, \mathbf{i}_M\}$ is the set of training images to learn the model parameters, $\mathbf{Y}^* = \{\mathbf{y}_1, \mathbf{y}_2, \cdots, \mathbf{y}_M\}$ is the collection of noisy embeddings in the training data, $p(\mathbf{y}|\mathbf{i}, \mathbf{I}^*, \mathbf{Y}^*)$ is the uncertainty of embedding estimation given the training images, and $p(\mathbf{i}|\mathbf{I}^*, \mathbf{Y}^*)$ is the probability density of face images in the underlying manifold of noiseless embeddings. In addition, we assume that the face of each identity has a deterministic function that maps the face to the corresponding noiseless embedding $\boldsymbol{\mu}$.

Suppose it is possible to capture a sufficient number of image samples of one identity to form a template $T = \{\mathbf{i}_1, \mathbf{i}_2, \cdots, \mathbf{i}_N\}$ , where $N$ is the number of images in the template. Since $N$ can be large in case of videos, the noiseless embedding $\boldsymbol{\mu}$ can be approximated by the expectation $\hat{E}(\mathbf{Y}^T)$:
\begin{equation}
    \boldsymbol{\mu} \approx \hat{E}(\mathbf{Y}^T) = \sum_{i=1}^N p(\mathbf{y}|\mathbf{I}^*, \mathbf{Y}^*) \mathbf{y}_i,
\end{equation}
where $\mathbf{Y}^T$ is the collection of noisy embeddings in the template $T$. However, estimating the probabilistic density of face embeddings while justifying various sources of noise in face representations is a very challenging task. An alternative solution for approximating the template representation is to estimate an adaptive scalar weight based on each feature vector (noisy embedding) and the approximated embedding of the template is the linear combination of the vectors based on the weights~\cite{yang2017neural}, ~\cite{liu2017quality}:
\begin{equation}
    \mathbf{r}^T = \sum_{i=1}^N f(\mathbf{y}_i) \mathbf{y}_i,
\end{equation}
where $\mathbf{r}^T$ is the template representation, and $f(\mathbf{y}_i)$ is the predicted weight for the feature vector of the $i^{th}$ image in the template. Although this kind of approach can reduce feature noise to some extent, the output weight is only inferred from the current feature vector. Considering a situation in which a video is captured under unconstrained conditions, e.g., surveillance cameras, most of the face frames are corrupted and only a small proportion has relatively high quality. Since the observed frame qualities are seriously unbalanced, such a weight estimation scheme may still be affected by observational error. In data assimilation for numerical weather prediction (NWP), it is important to build an observation bias correction scheme based on intricate (spatial and temporal) knowledge of the systemic errors~\cite{dee2005bias}. This inspires us to estimate quality weights based on context information by using all the images in a template to further diminish the impact of observational error in video clips. It has been shown that RNN is capable of capturing contextual cues by traversing the sequential slices along different directions~\cite{graves2008facial}. 

In this paper, we propose an RNN guided feature aggregation network which predicts a quality weight for each deep feature vector based on vectors in the same template. Similar to~\cite{gong2019video}, the proposed RNN-based model also generates different weights for each component (dimension) of the deep feature vectors. Hence each component of the template representation is:
\begin{equation}
    \mathbf{r}^T_j = \sum_{i=1}^N f(\mathbf{y}_{1j}, \mathbf{y}_{2j}, \cdots, \mathbf{y}_{Nj}) \mathbf{y}_{ij},
\end{equation}
where $\mathbf{r}^T_j$ is the $j^{th}$ component of the template representation, and $f(\mathbf{y}_{1j}, \mathbf{y}_{2j}, \cdots, \mathbf{y}_{Nj})$ is the predicted weight for the $j^{th}$ component of the feature vector of the $i^{th}$ image in the template. By using the context information, the weight value for each feature depends on other features in the template. Furthermore, the overall influence of poor quality components is diminished rather than enhanced by their large proportion in the template; frames with good quality can still dominate the final representation in spite of their lack of quantity.

\begin{table}[!h]
    \footnotesize
    \caption{Face Identification Accuracy (both for closed-set and open-set scenarios) of feature aggregation with and without context information on five protocols of IJB-S dataset.}
    \label{table:context_verify}
	\centering
	\begin{threeparttable}
        \scalebox{1.0}{
		\begin{tabularx}{1.0\linewidth}{c c c c c c c}
		\toprule
		\multirow{2}{*}{Test Protocol} & \multirow{2}{*}{Method} && \multicolumn{2}{c}{Closed-set (\%)} && Open-set (\%) \\
		\cline{4-5}
		& && Rank-1 & Rank-5 && 1 \% FPIR \\
		\midrule
        \multirow{2}{*}{SV\tnote{*} to still}          & No-Context && $49.57$ & $59.58$ && $16.48$ \\
	                                                    & Context && $\mathbf{52.14}$ & $\mathbf{61.46}$ && $\mathbf{20.59}$ \\
		\midrule
		\multirow{2}{*}{SV\tnote{*} to B\tnote{$\dagger$} }        & No-Context && $51.14$ & $61.43$ && $26.64$ \\
		                                                & Context && $\mathbf{53.81}$ & $\mathbf{63.23}$ && $\mathbf{30.39}$ \\
		\midrule
		Multi-view                                      & No-Context && $94.55$ & $99.01$ && $61.39$ \\
		SV\tnote{*} to B\tnote{$\dagger$}                         & Context && $\mathbf{97.52}$ & $\mathbf{99.01}$ && $\mathbf{74.26}$ \\
		\midrule
		\multirow{2}{*}{SV\tnote{*} to SV\tnote{*}}   & No-Context && $\mathbf{8.90}$ & $\mathbf{16.61}$ && $0.06$ \\
		                                                & Context && $8.85$ & $16.55$ && $\mathbf{0.09}$ \\
		\midrule
		UAV                                             & No-Context && $5.06$ & $11.39$ && $0.00$ \\
		SV\tnote{*} to B\tnote{$\dagger$}                         & Context && $\mathbf{6.33}$ & $\mathbf{13.92}$ && $0.00$ \\
		\bottomrule
		\end{tabularx}}
		
		\begin{tablenotes}\footnotesize
        \item[*] Surveillance videos
        \item[$\dagger$] Booking images (the full set of images, frontal and profile, taken of a subject at enrollment time)
        \end{tablenotes}
        
	\end{threeparttable}
\end{table}

% \begin{figure}
%     \centering
%     \includegraphics[scale=0.6]{figs/template_hist.pdf}
%     \caption{A histogram showing the percentage of low quality images in video templates of IJB-S dataset.}
%     \label{fig:template_hist}
% \end{figure}

\subsection{Illustration of Feature Aggregation}

In this section, we provide an example to demonstrate the efficacy of feature aggregation by adopting context information in a template. Using a pre-trained CNN embedding~\footnote{A 64-layer CNN~\cite{liu2017sphereface} trained on MS-Celeb-1M~\cite{guo2016msceleb} dataset.} for extracting facial features, we train a multilayer perceptron (MLP) with two fully connected layers to predict a quality indicator for each feature vector in a similar manner as~\cite{yang2017neural}. Then two types of aggregation strategies are considered:
\begin{itemize}\vspace{-0.6em}
    \item \textit{No-Context}: The quality scores are normalized with softmax function and the fused representation is obtained by aggregating all the feature vectors with the normalized scores.\vspace{-0.6em}
    \item \textit{Context}: The images are separated into two groups using k-means on the quality scores. The fused representation is obtained by only aggregating the group with higher quality.\vspace{-0.5em}
\end{itemize}
% that outputs a weight scalar for each feature vector extracted from a pre-trained CNN model. The weight is generated based on its corresponding feature vector, and no context information is taken into account. We use a triplet loss function~\cite{Schroff_2015_CVPR} to fine tune the parameters of the weight prediction layer, and the weight scalars in one template are then normalized with a softmax operator. Finally, the template representation is generated by aggregating the feature vectors with the normalized weights.

% In comparison, we take the context information into consideration by clustering the feature vectors and their weights into two groups using k-means algorithm. The first group consists of vectors with high weight values, i.e., good image quality, while the second comprises features with poor quality. 
% %We first compute the weighted average presentation in the second group, and take the mean value of the weights as the quality for the group presentation. Then we combine the new presentation with the feature vectors in the first group, and the re-normalize the weight scalars. The final template representation is then obtained from the aggregation of feature vectors with high weights and the pooled low quality features.
% We only take one single feature vector aggregated from the second group and fuse it with the vectors in group one. The final template is the weighted summation of these features.

Table~\ref{table:context_verify} reports the face recognition results of the above two aggregation strategies on IJB-S~\cite{Kalka2018IJBS} under five different protocols (Section~\ref{section:experiments}). It can be observed that the context information improves the face recognition performance on all five protocols, except the closed-set surveillance to surveillance protocol. For \textit{No-Context}, since most of the frames in IJB-S videos contain redundant and low quality faces (See Figure~\ref{fig:ijbs-s}), noisy embeddings in the template could still have a large impact because of their large proportion even when they are weighted by the quality scores. On the other hand, \textit{Context} is able to further improve the performance by simply dropping the potential redundant images (images predicted with lower quality). Note that for the surveillance to surveillance protocol, both probe set and gallery set are composed of noisy frames. Therefore, by removing low quality frames has little impact on the performance.
% Since we delete most of the video frames with low quality, some of which still contain useful information, the overall discrimination of the video presentation may be degraded in video to video scenarios.
% %We also show the histogram for the percentage of images with lower quality (images in the clustering with lower weight values) in video templates of IJB-S dataset (see Figure~\ref{fig:template_hist}). The histogram indicates 
% Note that more than one third of the video templates in IJB-S have over $50\%$ low quality images, suffering from severe observation bias (See Figure~\ref{fig:ijbs-s}). The context-aware pooling reduces the impact of poor images and enhances the overall discrimination of the representation.

\section{Approach}

\subsection{Overall Framework}

\begin{figure*}
    \centering
    \captionsetup{font=small}
    \includegraphics[width=1.0\linewidth]{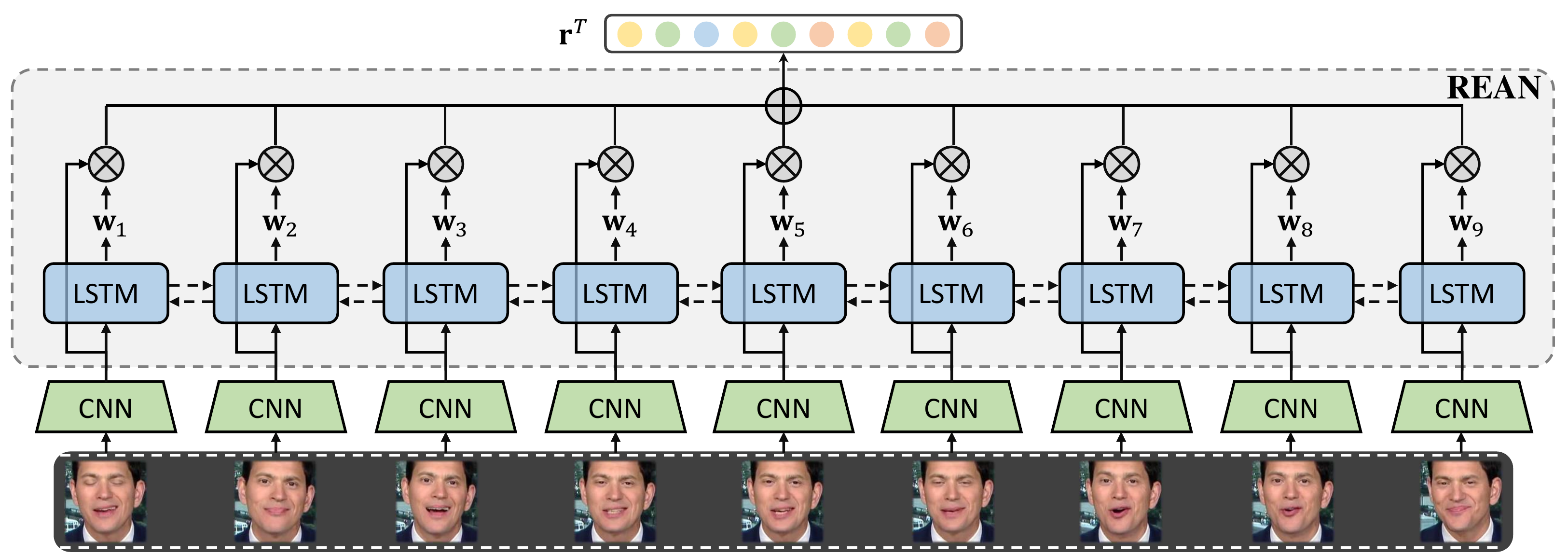}
    \caption{Overview of the proposed REAN. A set of face video frames are first accepted by a CNN model to extract deep face features. Then a bidirectional LSTM model is fed by these deep feature vectors to predict attention score for each feature. The network finally outputs a single feature vector as the face representation of the set for the recognition task.}
    \label{fig:overview}
\end{figure*}

The overall framework of the proposed context-aware attention network is presented in Figure~\ref{fig:overview}. A base CNN model is incorporated for extracting features from each face image and then the proposed Recurrent Embedding Aggregation Network (REAN)) is used to aggregate these features by considering the information in the whole video. The two models can be trained simultaneously in an end-to-end strategy, or separately. In this work,
% we train them one by one. The main reason is that existing video face datasets are rather small and are not practical for training a image-based face representation model. In particular,
we first train the base CNN on subjects in a large dataset of still face images (MS-Celeb-1M~\cite{guo2016msceleb}). Then it is used to extract the features from a video face dataset, UMDFaceVideo~\cite{klare2015pushing}, which is further used to train the REAN to adaptively predict quality weights for each deep feature vector. The features in a template are then aggregated with context-aware quality weights into a single compact feature vector.

\subsection{Context-aware Aggregation Module}

Let $\mathbf{F} = \{\mathbf{f}_1, \mathbf{f}_2, \cdots, \mathbf{f}_N\}$ be the CNN feature vectors of images in a template $T$, where each $\mathbf{f}_i$ is a $D$-dimensional vector and $N$ is the number of face images in the template. A hidden state $\mathbf{h}_t$ of LSTM at time step $t$ is computed based on the hidden state at time $t-1$, the input $\mathbf{f}_t$ and the cell state $\mathbf{C}_t$ at time $t$:
\begin{equation}
    \mathbf{h}_t = \sigma (\mathbf{W}_o[\mathbf{h}_{t-1}, \mathbf{f}_t] + b_o) \cdot tanh(\mathbf{C}_t),
\end{equation}
where $\sigma (\cdot)$ is the sigmoid function, $\mathbf{W}_o$ and $b_o$ are the parameters of the sigmoid gate. The quality vector for the $t^{th}$ feature vector in $\mathbf{F}$ is then inferred by the subsequent fully-connected layer: $\mathcal{H}(\mathbf{h}_t) = \mathbf{q}_t$, where $\mathbf{q}_t$ has the same dimensionality $D$ as $\mathbf{f}_t$. For the feature aggregation, we first employ a softmax operator to normalize all quality vectors in the template along each component. Specifically, given a set of quality vectors $\{\mathbf{q}_1, \mathbf{q}_2, \cdots, \mathbf{q}_N\}$, the $j^{th}$ component of the $t^{th}$ vector is normalized by:
\begin{equation}
    w_{tj} = \frac{\exp(q_{tj})}{\sum_{i=1}^N \exp(q_{ij})}
\end{equation}

The final template representation is the weighted mean vector of elements in $\mathbf{F}$:
\begin{equation}
\label{eq:template_feature}
    \mathbf{r}^T = \sum_{i=1}^N \mathbf{f}_i \odot \mathbf{w}_i,
\end{equation}
where $\odot$ denotes the element-wise multiplication, and $\mathbf{r}^T$ is a D-dimensional feature vector for template $T$.

\subsection{Network Training}

% The proposed REAN is trained on a video face dataset UMDFaceVideo~\cite{bansal2017s} using the features extracted from a pre-trained base CNN model. 
The architecture of REAN consists of a two-layer bi-directional LSTM network and a fully-connected layer. The fully-connected layer is needed to project the LSTM embeddings into the same dimension as the input feature vector. Note that the output of this module is the quality weights for each feature vector and the original CNN features is not modified. Finally, we aggregate the CNN features in the same template by leveraging the predicted weights for each component of the vector and use the weighted mean vector as the template representation. To optimize the weight prediction, we adopt a template-based triplet loss. The triplet comprises one anchor template, one positive template of the same subject as the anchor, and one negative template of a different identity. All the templates are randomly selected to form a mini-batch and average hard triplet is utilized. Here, the hard triplet means the non-zero loss triplets~\cite{hermans2017defense}. The loss function is formulated as:
\begin{equation}
   \mathcal{L}_{triplet} = \frac{1}{M} \sum_{i=1}^M {[{\|\mathbf{r}^{T_a}_i - \mathbf{r}^{T_p}_i\|}^2_2 - {\|\mathbf{r}^{T_a}_i - \mathbf{r}^{T_n}_i\|}^2_2 + \beta]}_+ ,
\end{equation}
where $M$ is the number of hard triplets in a mini-batch, and $\{\mathbf{r}^{T_a}_i, \mathbf{r}^{T_p}_i, \mathbf{r}^{T_n}_i\}$ stands for the $i^{th}$ triplet with anchor, positive, and negative template representations derived by Equation~\ref{eq:template_feature}. $[x]_+ = \max (0, x)$, and $\beta$ is the margin parameter.

%------------------------------------------------------------------------
\section{Experiments}
\label{section:experiments}

%-------------------------------------------------------------------------
\subsection{Datasets and Protocols}

We train REAN on UMDFaceVideo dataset~\cite{bansal2017s}, and evaluate it on three other video face datasets, the IARPA Janus Benchmark - Surveillance (IJB-S)~\cite{Kalka2018IJBS}, the YouTube Face dataset (YTF)~\cite{wolf2011face}, and the Point-and-Shoot Challenge dataset (PaSC)~\cite{beveridge2013challenge}.

\textbf{UMDFaceVideo} contains 3,735,476 annotated video frames extracted from a total of 22,075 videos for 3,107 subjects. The videos are collected from YouTube. The dataset is only used for training.

\textbf{IJB-S} is a surveillance video dataset collected by IARPA for face recognition system evaluation. The dataset is composed of 350 surveillance videos with 30 hours in total, 5,656 enrollment images, and 202 enrollment videos. The videos were captured under real-world environments, and can be used to simulate law enforcement and security applications. The dataset also contains several surveillance relevant protocols, including closed-set and open-set 1:N identification evaluations. The evaluations consist of five experiments: i) surveillance-to-still~\footnote{``Still'' stands for single frontal still images.}, ii) surveillance-to-booking~\footnote{The ``booking'' template comprises the full set of images captured of a single subject at enrollment time.}, iii) surveillance-to-surveillance, iv) multi-view surveillance-to-booking, and v) UAV~\footnote{UAV is a small fixed-wing unmanned aerial vehicle that was flown to collect images and videos.} surveillance-to-booking. IJB-S also includes a face detection protocol. Since our work mainly focuses on recognition and not detection, we only follow the five identification protocols and report the standard Identification Rate (IR) and open-set performance in terms of TPIR @ FPIR \footnote{True positive identification and false positive identification rate.}. Due to the poor quality of video frames in IJB-S, only 9 million out of 16 million faces could be detected. In our experiments, failure-to-enroll face images do not get utilized in template feature aggregation, and we use zero-vector as the template representation if no faces can be enrolled in the template.

\textbf{YTF} is a video face dataset that contains 3,425 videos of 1,595 different subjects. The average number of frames in YTF videos is 181. Unlike surveillance scenario of IJB-S, most videos in YTF are from video capture or posted on social media, where faces are more constrained and have higher quality. We report the 1:1 face verification rate of the given 5,000 video pairs in our experiments without fine-tuning.

\textbf{PaSC} is a video face dataset that contains 2,802 videos of 265 subjects, varying in viewpoints, sensor types, and distance to the camera, etc. The dataset consists of two subsets of videos captured by control and handheld cameras, respectively. Similar to IJB-S, no training set is provided by PaSC, so we 
% use our models trained on UMDFaceVideo dataset to 
directly evaluate the proposed approach on PaSC without further pre-processing or fine-tuning.

%-------------------------------------------------------------------------
\subsection{Implementation Details}
\textbf{Pre-processing}: All the face images in our experiments are automatically detected by a facial landmark detection algorithm, MTCNN~\cite{zhang2016joint}. Detected face regions are cropped from the original images and are resized into $112 \times 96$ after alignment by similarity transformation based on five facial landmarks~\footnote{Left eye, right eye, center of nose, left edge of mouth and right edge of mouth}.

\textbf{Training}: Our base CNN model is a 64-layer residual network~\cite{liu2017sphereface} trained on a clean version\footnote{\url{https://github.com/inlmouse/MS-Celeb-1M_WashList}} of MS-Celeb-1M dataset~\cite{guo2016msceleb} to learn a $512$-dimensional face representation of still images. The parameters of REAN are then trained on UMDFaceVideo~\cite{bansal2017s} with Adam optimizer, whose first momentum is set to $0.9$ and the second momentum is $0.999$. The margin of triplet loss is set to $3.0$. During training, we define one template as the frames in the same video of one subject and we fix the number of frames in one template. In particular, each mini-batch incorporates 384 templates that are randomly sampled from 128 subjects, 32 images per template. The model is trained for 20 epochs in total. We remove the subjects appear in the testing datasets and a small set of the training set is separated as validation set for tuning the  hyper-parameters. We conduct all the experiments on a Nvidia Geforce GTX 1080 Ti GPU and the average speed of feature extraction is 1ms per image. No further fine-tuning is employed.
%after training REAN on UMDFaceVideo.

%-------------------------------------------------------------------------
\subsection{Baseline}
We design three baseline experiments to compare with the proposed REAN.
\begin{itemize}
\vspace{-0.6em}
    \item \textit{\textbf{AvgPool}} simply applies the average pooling to the base CNN features to generate the template representation.\vspace{-0.6em}
    \item \textit{\textbf{LSTM}} is a two-layer LSTM network to predict the template representation directly without attention based feature aggregation. The output of the last cell is used as the representation.\vspace{-0.6em}
    \item \textit{\textbf{QualityPool}} is a two-layer fully-connected network with ReLU as the activation function in between. Similar to previous work~\cite{yang2017neural,liu2017quality,gong2019video}, the model also predicts quality weights for each feature vector and takes the weighted summation of all vectors as the template representation.\vspace{-0.6em}
\end{itemize}

%-------------------------------------------------------------------------
\subsection{Qualitative Analysis of IJB-S}

\begin{figure*}
    \centering
    \captionsetup{font=small}
    \subfloat[Example from IJB-S]{\includegraphics[width=1.0\linewidth]{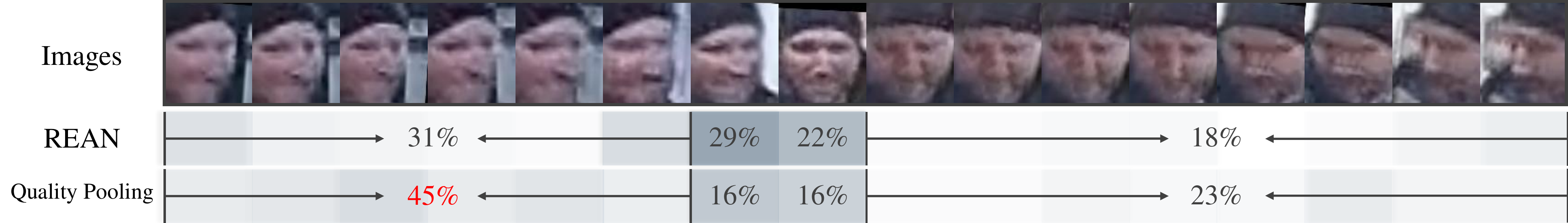}}\\
    \subfloat[Example from PaSC]{\includegraphics[width=1.0\linewidth]{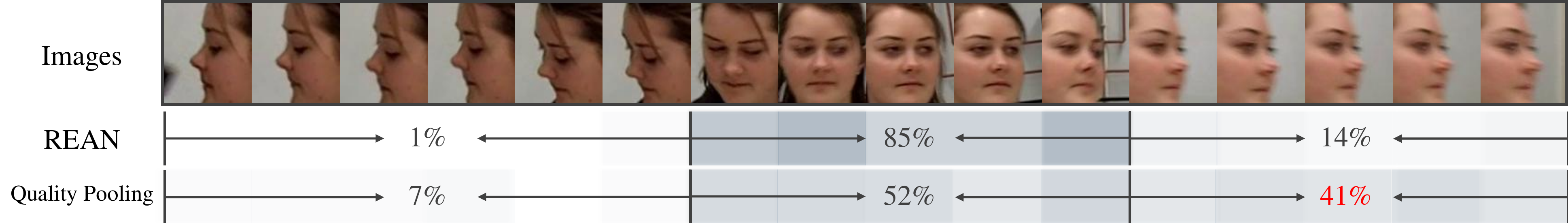}}
    \caption{Attention distribution of REAN and quality-aware pooling on two video sequences from IJB-S and PaSC. The summation of the weight percentage is given below each clustering of the frames. The darker the blue box is, the higher the attention weight of the frame above. Although ``Quality Pooling'' is able to assign larger weights to higher quality images, it can be distracted by redundant low-quality images if there is a large number of them (red numbers). In comparison, REAN is able to keep focused on high quality frames by leveraging the context information. The attention here is computed by averaging across all components of the weight vector.}
    \label{fig:pooling}
\end{figure*}

To evaluate the effect of context-aware pooling by REAN, we visualize the attention distribution on two example video sequences from IJB-S and PaSC. The two sequences are composed by randomly sampling $16$ images from the original video. Then the two models, \textit{QualityPool} and REAN, are used to compute the attention for the images in the sequence. For visualization purpose, the attention vector (for different components) of each image is averaged into one scalar. Two example results are shown in Figure~\ref{fig:pooling}. \textit{QualityPool} can effectively predict the quality of the given image, but without context information, it is vulnerable against redundant low-quality images,~\ie when the number of low quality images is high, it could be easily distracted. The first $6$ images in the IJB-S video are nearly identical images, but overall they have even a larger weight than the two higher quality faces. Similarly, in the PaSC case, the last $5$ images almost contain the same information, but overall they gain $41\%$ attention. In comparison, REAN is more robust against redundancy. 

%-------------------------------------------------------------------------
\subsection{Quantitative Analysis on IJB-S}

\begin{table*}[!h]
    \centering
    \footnotesize
    \caption{Performance comparisons on IJB-S dataset.}
    \label{table:ijbs}
		\centering
		\begin{tabularx}{0.9\linewidth}{X c c c c c c c c}
		\toprule
		\multirow{2}{*}{Test Name} & \multirow{2}{*}{Method} && \multicolumn{3}{c}{Closed-set (\%)} && \multicolumn{2}{c}{Open-set (\%)} \\
		\cline{4-6} \cline{8-9}
		& && Rank-1 & Rank-5 & Rank-10 && 1 \% FPIR & 10 \% FPIR \\
		\midrule
        \multirow{5}{*}{Surveillance-to-still}          & C-FAN~\cite{gong2019video} && $50.82$ & $61.16$ & $64.95$ && $16.44$ & $24.19$ \\
                                                        & \textit{AvgPool} && $50.80$ & $59.60$ & $64.86$ && $11.60$ & $20.84$ \\
                                                        & \textit{LSTM} && $2.90$ & $17.46$ & $38.10$ && $0.12$ & $2.80$ \\
	                                                    & \textit{QualityPool} && $51.61
	                                                    $ & $62.78$ & $66.39$ && $17.33$ & $23.91$ \\
	                                                    & REAN && $\mathbf{57.59}$ & $\mathbf{64.07}$ & $\mathbf{68.39}$ && $\mathbf{20.06}$ & $\mathbf{28.44}$ \\
		\midrule
		\multirow{5}{*}{Surveillance-to-booking}        & C-FAN~\cite{gong2019video} && $53.04$ & $62.67$ & $66.35$ && $27.40$ & $29.70$ \\
		                                                & \textit{AvgPool} && $50.82$ & $59.73$ & $64.25$ && $19.19$ & $25.43$ \\
		                                                & \textit{LSTM} && $4.49$ & $16.40$ & $32.81$ && $1.31$ & $15.97$ \\
		                                                & \textit{QualityPool} && $52.66$ & $62.95$ & $65.86$ && $25.60$ & $31.83$ \\
		                                                & REAN && $\mathbf{58.52}$ & $\mathbf{65.21}$ & $\mathbf{68.62}$ && $\mathbf{31.25}$ & $\mathbf{33.10}$ \\
		\midrule
		                                            & C-FAN~\cite{gong2019video} && $96.04$ & $99.50$ & $99.50$ && $70.79$ & $85.15$ \\
		Multi-view                                  & \textit{AvgPool} && $96.53$ & $98.51$ & $99.01$ && $66.33$ & $82.17$ \\
		                                            & \textit{LSTM} && $4.95$ & $19.80$ & $39.61$ && $3.21$ & $23.56$ \\
	    Surveillance-to-booking                     & \textit{QualityPool} && $97.03$ & $99.50$ & $99.50$ && $75.62$ & $86.21$ \\
	                                                & REAN && $\mathbf{98.02}$ & $\mathbf{99.50}$ & $\mathbf{99.50}$ && $\mathbf{76.24}$ & $\mathbf{93.07}$ \\
		\midrule
		\multirow{5}{*}{Surveillance-to-Surveillance}   & C-FAN~\cite{gong2019video} && $10.05$ & $17.55$ & $21.06$ && $0.11$ & $0.68$ \\
		                                                & \textit{AvgPool} && $7.71$ & $14.34$ & $18.95$ && $0.08$ & $0.57$ \\
		                                                & \textit{LSTM} && $11.13$ & $21.07$ & $23.52$ && $0.08$ & $0.20$ \\
		                                                & \textit{QualityPool} && $5.69$ & $9.43$ & $13.00$ && $0.09$ & $0.34$ \\
		                                                & REAN && $\mathbf{21.96}$ & $\mathbf{33.79}$ & $\mathbf{36.30}$ && $\mathbf{0.21}$ & $\mathbf{0.97}$ \\
		\midrule
		                                                & C-FAN~\cite{gong2019video} && $7.59$ & $\mathbf{12.66}$ & $20.25$ && $0.00$ & $0.00$ \\
	    UAV                                             & \textit{AvgPool} && $2.53$ & $6.33$ & $7.59$ && $0.00$ & $0.00$ \\
	                                                    & \textit{LSTM} && $1.30$ & $2.66$ & $6.58$ && $0.00$ & $0.00$ \\
		Surveillance-to-booking                         & \textit{QualityPool} && $7.59$ & $10.85$ & $13.92$ && $0.00$ & $1.04$ \\
		                                                & REAN && $\mathbf{8.06}$ & $11.39$ & $\mathbf{23.92}$ && $\mathbf{3.13}$ & $\mathbf{3.13}$ \\
		\bottomrule
		\end{tabularx}
\end{table*}

Table~\ref{table:ijbs} reports identification results on IJB-S dataset. One of the comparisons is between the proposed method and previous work on IJB-S~\cite{gong2019video}. It is clear that the proposed approach achieves the best performance on most of the five protocols. In particular, the proposed approach outperforms C-FAN by $11.91\%$ and $15.24\%$ on the closed-set surveillance-to-surveillance protocol at rank 1 and rank 5, respectively. Moreover, with the proposed technique, the open-set protocol results are also improved by $3.62\%$, $3.85\%$, and $5.45\%$ on surveillance-still, surveillance-to-booking, and multi-view surveillance-to-booking at $1\%$ FPIR, respectively. It is worth noting that REAN is trained on video frames data with temporal information. However, the booking images in IJB-S are still face images with high quality, on which the quality prediction may not be as accurate as surveillance frames. Therefore, for booking images, we use average pooling instead of quality aggregation. In comparison with the three baseline methods, REAN achieves higher identification rates than \textit{LSTM} on all five protocols. Obviously, the \textit{LSTM} over fits to the video frames in UMDFaceVideo and is not able to generalize to the booking images. Since the range of co-domain of \textit{LSTM} is larger than that of REAN, the distribution of the output vector from \textit{LSTM} can be very different from original feature vector, so the matching between surveillance videos and still images leads to poor performance. \textit{QualityPool} achieves similar performance as C-FAN, since both of the approaches use component-wise attention scheme. Both \textit{QualityPool} and the proposed REAN outperform average pooling of the base CNN features.

%-------------------------------------------------------------------------
\subsection{Performance Comparison on YTF and PaSC}

\begin{table}[t]
    \centering
    \footnotesize
    \caption{Verification Performance on YouTube Face dataset, compared with baseline methods and other state-of-the-art methods.}
    \label{table:ytf}
    \scalebox{1.0}{
    \begin{tabularx}{\linewidth}{X c X c}
        \toprule
        Method & Accuracy (\%) & Method & Accuracy (\%)\\
        \midrule  
        EigenPEP~\cite{li2014eigen} & $84.8 \pm 1.4$ & DeepFace \cite{taigman2014deepface} & $91.4 \pm 1.1$ \\
        DeepID2+~\cite{sun2015deeply} & $93.2 \pm 0.2$ & C-FAN~\cite{gong2019video} & $96.50 \pm 0.90$ \\
        FaceNet~\cite{Schroff_2015_CVPR} & $95.52 \pm 0.06$ & DAN~\cite{rao2017learning} & $94.28 \pm 0.69$ \\
        NAN~\cite{yang2017neural} & $95.72 \pm 0.64$ & QAN~\cite{liu2017quality} & $96.17 \pm 0.09$ \\
        \midrule
        \textit{AvgPool} & $96.24 \pm 0.96$ & \textit{LSTM} & $60.00 \pm 2.81$ \\
        \textit{QualityPool} & $96.38 \pm 0.95$ & REAN & $\mathbf{96.60} \pm \mathbf{1.00}$\\
        \bottomrule
    \end{tabularx}}
\end{table}

\begin{table}[t]
    \centering
    \footnotesize
    \caption{Comparisons of the verification rate (\%) with the state-of-the-art methods on PaSC at a false accept rate (FAR) of 0.01.}
    \label{table:pasc}
    \begin{tabularx}{0.7\linewidth}{X c c}
    \toprule
    Method & Control & Handheld \\
    \midrule
    DeepO2P~\cite{mohagheghian2016application} & $68.76$ & $60.14$ \\
    SPDNet~\cite{huang2017riemannian} & $80.12$ & $72.83$ \\
    GrNet~\cite{huang2018building} & $80.52$ & $72.76$ \\
    Rao~\etal~\cite{rao2017attention} & $95.67$ & $93.78$ \\
    TBE-CNN~\cite{ding2018trunk} & $95.83$ & $94.80$ \\
    TBE-CNN + BN~\cite{ding2018trunk} & $97.80$ & $96.12$ \\
    \midrule
    \textit{AvgPool} & $91.27$ & $74.30$ \\
    \textit{LSTM} & $3.07$ & $1.28$ \\
    \textit{QualityPool} & $96.48$ & $92.39$ \\
    REAN & $96.52$ & $94.63$ \\
    \bottomrule
    \end{tabularx}
\end{table}

Table~\ref{table:ytf} reports the face verification performance of the proposed method and other state-of-the-art methods on YTF dataset. REAN outperforms the three baseline methods in terms of average verification accuracy. The performance of REAN is slightly higher than previous approaches such as NAN~\cite{yang2017neural} and C-FAN~\cite{gong2019video}. Since YouTube Face videos are not captured by typical surveillance cameras, instead, most of them are recorded by professional photographers. As a result, they are free from very low quality frames. For this reason, the proposed context-aware attention network does not have obvious advantages over other state-of-the-art approaches. 

Table~\ref{table:pasc} reports the verification results on PaSC dataset. In comparison with YTF, PaSC is more challenging since the faces in the dataset have full pose variations. By comparing the proposed REAN with the three baseline models, we can observe that combining context information with attention aggregation is capable of improving the discriminative power of video face representations. We can also observe that the proposed approach achieves comparable performance compared to other state-of-the-art methods.

%------------------------------------------------------------------------
\section{Conclusions}
This paper addressed video face recognition for unconstrained surveillance systems. We propose a Recurrent Embedding Aggregation Network (REAN) that adaptively predicts context-aware quality vectors for each deep feature vector extracted by CNN face model. And each face in a video can perform efficient recognition algorithm by using a compact deep feature vector aggregated from REAN under the quality attention scheme. Experiments have been conducted on three video face datasets, i.e., IJB-S, YTF, and PaSC. Based on the recognition results, we empirically demonstrate that the attention values provided by the proposed REAN is able to maintain the discriminative features while discard the noisy features by leveraging the context information in the video learned by LSTM. Our method shows clear advantages on video face benchmarks, including surveillance videos.

{\small
\bibliographystyle{ieee}
\bibliography{egbib}
}

\end{document}